\documentclass{article}

     \usepackage[nonatbib, preprint]{neurips_2020}

\usepackage[utf8]{inputenc} 
\usepackage[T1]{fontenc}    
\usepackage{hyperref}       
\usepackage{url}            
\usepackage{booktabs}       
\usepackage{amsfonts}       
\usepackage{nicefrac}       
\usepackage{microtype}      

\usepackage{amsmath, amsfonts}
\usepackage{siunitx}
\usepackage[ruled,vlined]{algorithm2e}

\usepackage{booktabs}
\usepackage{epsfig}
\usepackage{etoolbox}
\usepackage{float}

\usepackage{array}
\newcolumntype{C}[1]{>{\centering\arraybackslash}p{#1}}

\usepackage{caption}
\usepackage{subcaption}
\def\E{{\rm E}}

\usepackage{xcolor}

\usepackage{multirow}

\title{Learning Latent Space Energy-Based Prior Model for Molecule Generation}

\author{
Bo Pang$^1$ \hskip1em Tian Han$^2$ \hskip1em Ying Nian Wu$^1$ \\
$^1$University of California, Los Angeles \hskip1em $^2$Stevens Institute of Technology
\\ {\small \texttt{bopang@ucla.edu}} \hskip1em {\small \texttt{than6@stevens.edu}} \hskip1em {\small \texttt{ywu@stat.ucla.edu}}
}

\begin{document}

\maketitle

\begin{abstract}
Deep generative models have recently been applied to molecule design. If the molecules are encoded in linear SMILES strings, modeling becomes convenient. However,  models relying on string representations tend to generate invalid samples and duplicates. Prior work addressed these issues by building models on chemically-valid fragments or explicitly enforcing chemical rules in the generation process. We argue that an expressive model is sufficient to implicitly and automatically learn the complicated chemical rules from the data, even if molecules are encoded in simple character-level SMILES strings. We propose to learn latent space energy-based prior model with SMILES representation for molecule modeling. Our experiments show that our method is able to generate molecules with validity and uniqueness competitive with state-of-the-art models. Interestingly, generated molecules have structural and chemical features whose distributions almost perfectly match those of the real molecules. 
\end{abstract}

\section{Introduction}
Designing molecules with desired properties is of vital importance in applications such as drug design and material science. Molecules are in the form of graphs. It is hence challenging to search for desirable ones in the molecule space. Recently, deep generative models have been applied to molecule modeling \cite{gomez2018automatic, kusner2017grammar, simonovsky2018graphvae, shi2020graphaf, podda2020deep}. Most methods adopt Variational Autoencoder (VAE) model \cite{kingma2013auto}. It embeds molecules into a continuous latent space, allowing for more efficient optimization, and then decodes the latent vector to a molecule, enabling new molecule generation.  

In molecule modeling, two types of representations are widely used. One is simplified molecular input line entry systems (SMILES) \cite{weininger1988smiles} with which a molecule graph is linearized into a string consisting of characters that represent atoms and bonds. With this representation, an autoregressive model can be utilized to capture the chemical rules among atoms and bonds. The same model is widely used and called language model (LM) in natural language processing. Following \cite{podda2020deep}, we call models adopting this representation as LM-based models. Another representation works directly with the graph where nodes and edges represent atoms and bonds respectively. Graph allows for explicitly encoding and directly enforcing chemical laws. To guarantee validity of generated molecules, many graph-based models \cite{liu2018constrained, samanta2019nevae, shi2020graphaf} sequentially generate atoms (nodes) and bonds (edges), continuously check if the generated elements satisfy valency rules. Graph-based models are however more complicated and less efficient to train and sample from, compared to LM-based models.  

Despite the simplicity and efficiency of LM-based models, they often produce invalid samples and duplicates. The recent work of \cite{podda2020deep} proposed FragmentVAE and argued that LM-based models can produce samples with perfect validity and uniqueness.  Fragments are small-weight and chemically sound compounds, and FragmentVAE uses fragments instead of atoms as basic elements in molecule generation. To enhance uniqueness, FragmentVAE replaces infrequent fragments in generated molecules by new fragments that are uniformly sampled from a pool of infrequent fragments. These techniques make the SMILES-fragment-based model competitive with the state-of-the-art graph-based models.

Instead of redesigning molecule representation or resorting to more complicated graph models, we argue that an expressive model is sufficient to capture the complicated chemical rules implicitly and generate valid and unique molecules, even with the character-level SMILES representation instead of fragment-level representation. Previous VAE-based methods rely on a generator network to map a prior distribution to be close to the data distribution and assume the prior to be a simple isotropic Gaussian distribution. Although a neural network generator is highly expressive, the assumption on the prior may cause ineffective learning of the model, which might explain why previous methods fail to generate valid and unique molecules without explicitly enforcing chemical rules. In this article, we propose to learn a latent space energy-based prior model \cite{pang2020learning} in addition to the generator network from observed molecules. Specifically, the prior model is an energy-based correction of the isotropic Gaussian distribution and the correction is learned from empirical data. Such a prior model improves the expressivity of the generator model.  Our experiments demonstrate that our method is able to generate valid and unique samples, with the performance on par with the state-of-the-art models. Interestingly, we observe that the generated samples show structural and chemical properties (e.g., solubility, drug-likeness) that closely resemble the ground truth molecules.

\section{Methods}

\subsection{Model}

Let $x \in \mathbb{R}^D$ be an observed molecule such as represented in SMILES strings. Let $z \in \mathbb{R}^d$ be the latent variables, where $D \gg d$. Consider the following model, 
\begin{align}
z \sim p_{\alpha}(z), \quad x \sim p_\beta(x|z),
\end{align}
where $p_{\alpha}(z)$ is the prior model with parameters $\alpha$, $p_\beta(x|z)$ is the top-down generative model with parameters $\beta$. In VAE, the prior is simply assumed to be an isotropic Gaussian distribution. In our model, $p_{\alpha}(z)$ is formulated as an energy-based model or a Gibbs distribution,
\begin{align}
p_{\alpha}(z) = \frac{1}{Z(\alpha)} \exp(f_{\alpha}(z)) p_0(z). \label{eq:prior}
\end{align}   
where $p_0(z)$ is a reference distribution, assumed to be isotropic Gaussian as in VAE. $f_{\alpha}(z)$ is the negative energy and is parameterized by a small multi-layer perceptron with parameters $\alpha$. $Z(\alpha) = \int \exp(f_\alpha (z)) p_0(z) dz = \E_{p_0}[\exp(f_\alpha(z))]$ is the normalizing constant or partition function. 

The generative model, $p_{\beta}(x|z)$, is a conditional autoregressive model,
\begin{align} 
    p_\beta(x|z) = \prod_{t=1}^T p_\beta(x^{(t)}|x^{(1)}, ..., x^{(t-1)}, z) 
\end{align} 
which is parameterized by a simple recurrent network with parameters $\beta$ and $x^{(t)}$ indicates a one-hot encoded SMILES string. 

It is worth pointing out the simplicity of the generative model of our method considering that those in prior work involve complicated graph search algorithm or alternating generation of atoms and bonds with multiple networks.

\subsection{Learning Algorithm} 

Suppose we observe training examples $(x_i, i = 1, ..., n)$. The log-likelihood function is 
\begin{align} 
  L( \theta) = \sum_{i=1}^{n} \log p_\theta(x_i). \label{eq:loglik}
\end{align}
where $\theta = (\alpha, \beta)$. The learning gradient can be calculated according to 
\begin{align} 
   \nabla_\theta  \log p_\theta(x) & = \E_{p_\theta(z|x)} \left[ \nabla_\theta \log p_\theta(x, z) \right] 
     = \E_{p_\theta(z|x)} \left[ \nabla_\theta (\log p_\alpha(z) + \log p_\beta(x|z)) \right].
\end{align}
For the prior model, 
\(
    \nabla_\alpha \log p_\alpha(z)  =  \nabla_\alpha f_\alpha(z) - \E_{p_\alpha(z)}[ \nabla_\alpha f_\alpha(z)]. 
\)
Thus the learning gradient for an example $x$ is
\begin{align} 
  \delta_\alpha(x) =   \nabla_\alpha \log p_\theta(x) = \E_{p_\theta(z|x)}[\nabla_\alpha f_\alpha(z)] - \E_{p_\alpha(z)} [\nabla_\alpha f_\alpha(z)]. \label{eq:alpha}
\end{align}
$\alpha$ is updated based on the difference between $z$ inferred from empirical observation $x$, and $z$ sampled from the current prior model.  

For the generative model,
\begin{align} 
\label{eq:beta}
\begin{split}
  \delta_\beta(x) =  \nabla_\beta \log p_\theta(x)  = \E_{p_\theta(z|x)} [\nabla_\beta \log p_{\beta}(x|z)],
\end{split}
\end{align} 
where $\sum_{t=1}^T \log p_\beta(x^{(t)}|x^{(1)}, ..., x^{(t-1)}, z)$ for text modeling which is about the reconstruction error. 

Expectations in (\ref{eq:alpha}) and (\ref{eq:beta}) require MCMC sampling of the prior model $p_\alpha(z)$ and the posterior distribution $p_\theta(z|x)$.  Instead of learning a separate network for approximate inference, we follow \cite{pang2020learning} and use Langevin dynamics for short run MCMC which iterates: 
\begin{align} 
z_0 \sim p_0(z), \; z_{k+1} = z_k + s \nabla_z \log \pi(z_k) + \sqrt{2s} \epsilon_k, \; k = 1, ..., K. 
\label{eq:Langevin}
\end{align}
where we initialize the dynamics from the fixed prior distribution of $z$, i.e., $p(z)\sim  {\rm N}(0, I_d)$ and $\epsilon_k \sim {\rm N}(0, I_d)$ is the Gaussian white noise. $\pi(z)$ can be either $p_\alpha(z)$ or $p_\theta(z|x)$. In either case, $\nabla_z \log \pi(z)$ can be efficiently computed by back-propagation. The dynamics runs a fixed number of $K$ steps with step size $s$.Denote the distribution of $z_K$ to be $\tilde{\pi}(z)$. As shown in \cite{cover2012elements}, the Kullback-Leibler divergence $D_{KL}(\tilde{\pi} \| \pi)$ decreases to zero monotonically as $K \rightarrow \infty$. 

Specifically, denote the distribution of $z_K$ to be $\tilde{p}_\alpha(z)$ if the target $\pi(z) = p_\alpha(z)$, and denote the distribution of $z_K$ to be $\tilde{p}_\theta(z|x)$ if $\pi(z) = p_\theta(z|x)$. The learning gradients in equations (\ref{eq:alpha}) and (\ref{eq:beta}) are modified to 
\begin{align} 
  &  \tilde{\delta}_\alpha(x) = \E_{\tilde{p}_\theta(z|x)}[\nabla_\alpha f_\alpha(z)] - \E_{\tilde{p}_\alpha(z)} [\nabla_\alpha f_\alpha(z)], \label{eq:alpha1}\\
  &  \tilde{\delta}_\beta(x)  = \E_{\tilde{p}_\theta(z|x)} [\nabla_\beta \log p_\beta(x|z)].  \label{eq:beta1}
\end{align} 
We then update $\alpha$ and $\beta$ based on (\ref{eq:alpha1}) and (\ref{eq:beta1}), where the expectations can be approximated by Monte Carlo samples. See \cite{pang2020learning} for theoretical foundation of the resulting learning algorithm. The short-run MCMC is efficient and mixes well in latent space due to the relative low-dimensionality of the latent space.

\section{Experiments}
\begin{figure}[H]
     \centering
     \begin{subfigure}[b]{0.49\textwidth}
         \centering
         \includegraphics[width=0.8\textwidth]{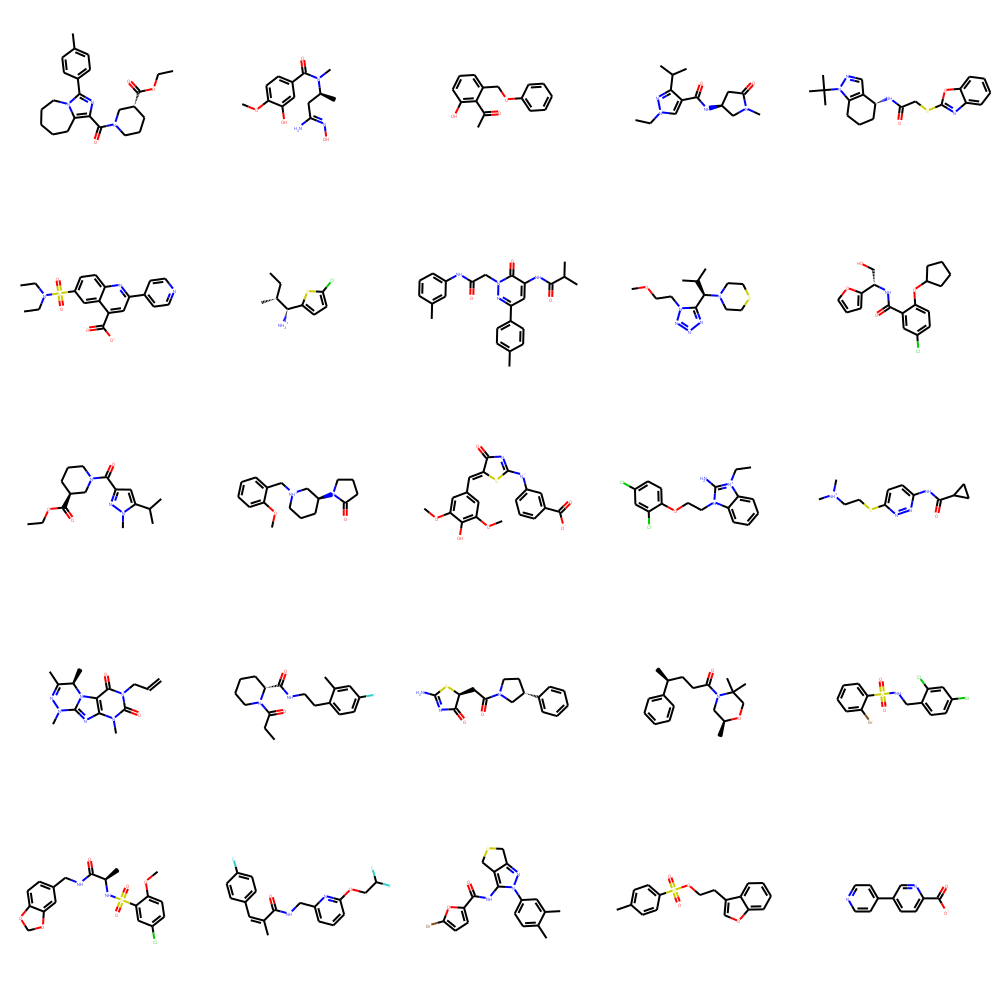}
         \caption{\small ZINC}
         \label{fig:zinc-samples}
     \end{subfigure}
     \hfill
     \begin{subfigure}[b]{0.49\textwidth}
         \centering
         \includegraphics[width=0.8\textwidth]{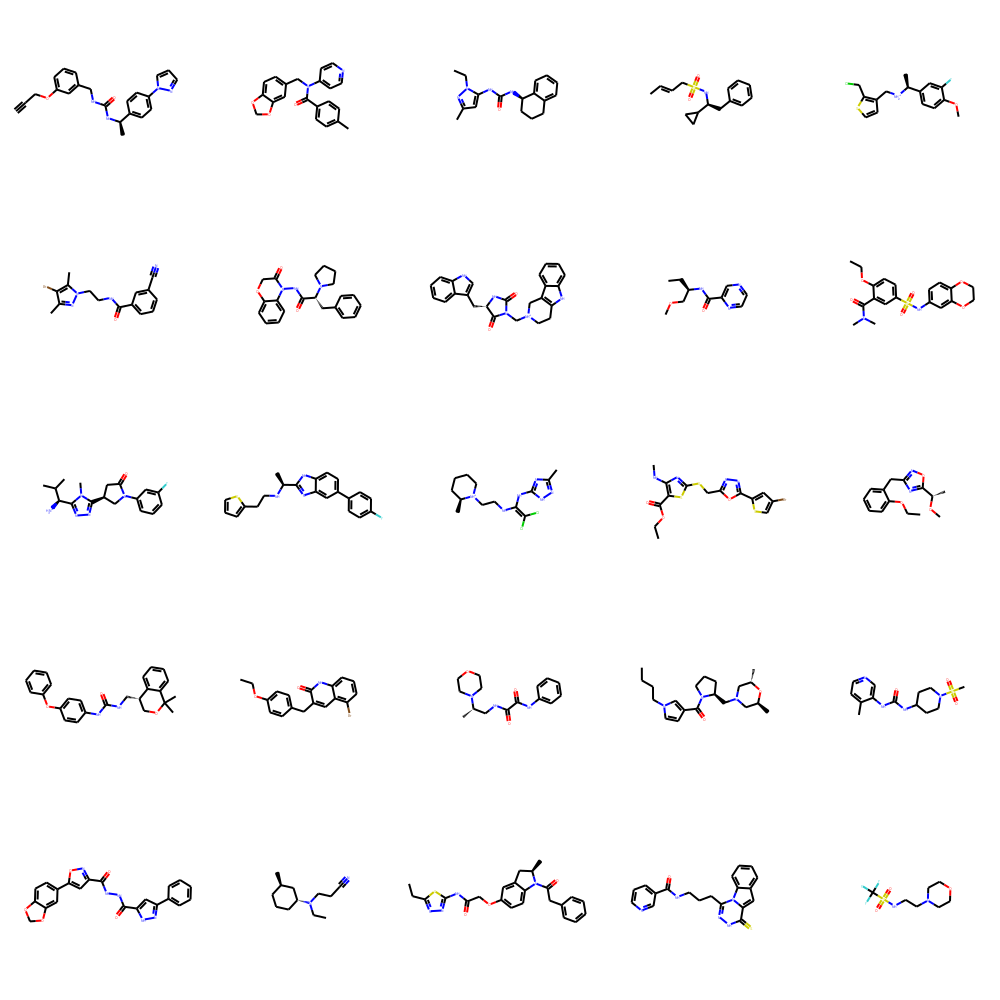}
         \caption{\small Generated}
         \label{fig:generated-samples}
     \end{subfigure}
     \vspace{2mm}
        \caption{\small Sample molecules taken from the ZINC dataset (a) and generated by our model (b).}
        \label{fig:samples}
\end{figure}

A standard molecule dataset, ZINC \cite{irwin2012zinc}, is used in our experiments. The latent space dimension is 32. The latent space energy-based model is implemented with a three-layer MLP with hidden dimension 200. The generator is a single layer LSTM with a hidden dimension of 1024 and the embedding dimension is 512. Figure~\ref{fig:samples} shows sample molecules generated from the data and randomly generated from our model. 

\subsection{Validity, novelty, and uniqueness}
We evaluate our model with three commonly used metrics: 1) validity, the percentage of valid molecules among all the generated ones; 2) novelty, the percentage of generated molecules not appearing in training set; 3) uniqueness, the percentage of unique ones among all the generated molecules. All metrics are computed based on 10,000 randomly generated molecules. Our model greatly improve previous LM-based models on validity and uniqueness and are competitive with fragment-based model and graph-based models using valency check. It is interesting to notice that the state-of-the-art graph-based models such as GCPN \cite{you2018graph} and GraphAF \cite{shi2020graphaf}, generate molecules with low validity rates if valency check is not applied. It appears that the graph-based models do not capture the chemical rules but instead strongly relies on explicit constraints. In contrast, our model is able to automatically learn the rules from the data. 

\begin{table}[H]
\scriptsize
\begin{center}
\begin{tabular}{lccccc}
    \toprule
    \textbf{Model} & \textbf{Model Family} & \textbf{Validity w/ check} & \textbf{Validity w/o check} & \textbf{Novelty} & \textbf{Uniqueness}\\
    \midrule
    GraphVAE (Simonovsky et al., 2018)    & Graph & 0.140 & - & 1.000 & 0.316\\
    CGVAE (Liu et al., 2018)      & Graph & 1.000 & - & 1.000 & 0.998\\
    GCPN (You et al., 2018)     & Graph & 1.000 & 0.200 & 1.000 & 1.000\\
    NeVAE (Samanta et al., 2019)     & Graph & 1.000 & - & 0.999 & 1.000\\
    MRNN (Popova et al., 2019)     & Graph & 1.000 & 0.650 & 1.000 & 0.999\\
    GraphNVP (Madhawa et al., 2019)  & Graph &  0.426 & - & 1.000 & 0.948 \\
    GraphAF (Shi et al., 2020)       & Graph & 1.000 & 0.680 & 1.000 & 0.991 \\
    \midrule
    ChemVAE (Gomez-Bombarelli et al., 2018)     & LM    & 0.170 & - & 0.980 & 0.310\\
    GrammarVAE (Kusner et al., 2017)  & LM    & 0.310 & - & 1.000 & 0.108\\
    SDVAE (Dai et al., 2018)       & LM    & 0.435 & - & - & -\\
    FragmentVAE (Podda et al., 2020)     & LM    & \textbf{1.000} & -  & 0.995 & 0.998\\
    \textbf{Ours}  & LM    & 0.955 & - & \textbf{1.000} & \textbf{1.000}\\
    \bottomrule
\end{tabular}
\vspace{1mm}
\caption{\small Performance obtained by our model against LM-based and graph-based baselines.}\label{tab:results}

\end{center}
\end{table}

\subsection{Molecular properties of samples}
\begin{figure}[H]
\begin{center}
	\begin{subfigure}{.32\textwidth}
		\centering
		\includegraphics[width=1.0\linewidth]{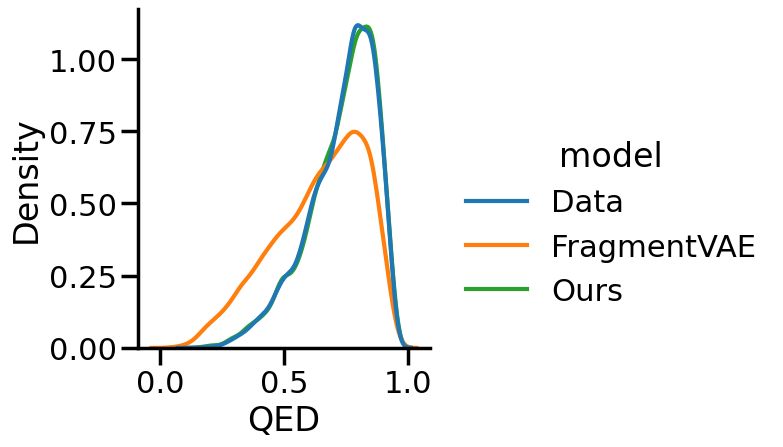}
		\label{fig:qed}
	\end{subfigure}
	\begin{subfigure}{.32\textwidth}
		\centering
		\includegraphics[width=1\linewidth]{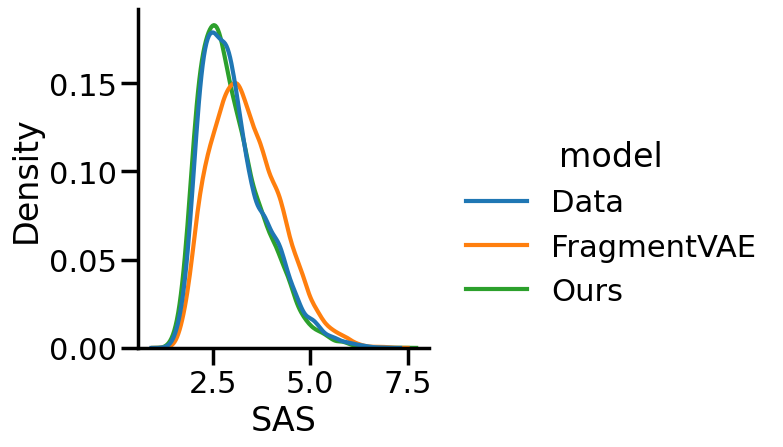}
		\label{fig:sas}
	\end{subfigure}
	\begin{subfigure}{.32\textwidth}
		\centering
		\includegraphics[width=\linewidth]{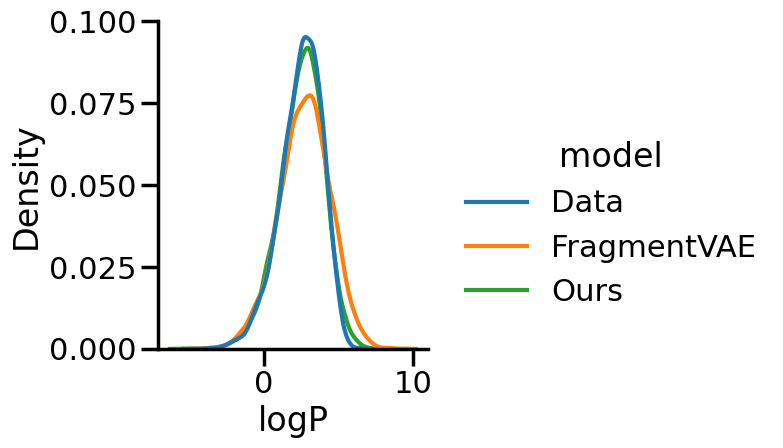}
		\label{fig:logp}
	\end{subfigure}
	\caption{\small Distributions of molecular properties of data and 10,000 random samples from FragmentVAE and our model.}
	\label{fig:kde}
	\end{center}
\end{figure}

If a model distribution matches the data distribution well, marginal distributions of any statistics would also match. Three properties are critical for molecule modeling, especially in \textit{de novo} drug design: 1) octanol/water partition coefficient (logP) which measures solubility; 2) quantitative estimate of drug-likeness (QED); 3) synthetic accessiblity score (SAS) which measures ease of synthesis. Each property can be viewed a statistic of the molecule data. In Figure~\ref{fig:kde}, we compare the distributions of the three properties based on 10,000 samples from the data and our model. The distributions based on FragmentVAE are also included for a reference. It is clear that our model produces distributions close to data property distributions, even though there is not any explicit supervision given for learning the three molecular properties. Also, our model evidently improve over FragmentVAE in this regard.

\section{Conclusion}
This work proposes to jointly learn a latent space energy-based prior model and a simple autoregressive generator for molecule modeling. Our approach yields a simple yet highly expressive model. The learned model generates valid and unique molecules with character-level SMILES representation. Key chemical properties of the generated samples closely resemble those of the data on a distribution level. These results provide strong evidence that the proposed model is able to automatically learn complicated chemical rules implicitly from the data.

\bibliographystyle{ieee_fullname}
\bibliography{main}

\begin{thebibliography}{10}\itemsep=-1pt

\bibitem{cover2012elements}
Thomas~M. Cover and Joy~A. Thomas.
\newblock {\em Elements of information theory {(2.} ed.)}.
\newblock Wiley, 2006.

\bibitem{gomez2018automatic}
Rafael G{\'o}mez-Bombarelli, Jennifer~N Wei, David Duvenaud, Jos{\'e}~Miguel
  Hern{\'a}ndez-Lobato, Benjam{\'\i}n S{\'a}nchez-Lengeling, Dennis Sheberla,
  Jorge Aguilera-Iparraguirre, Timothy~D Hirzel, Ryan~P Adams, and Al{\'a}n
  Aspuru-Guzik.
\newblock Automatic chemical design using a data-driven continuous
  representation of molecules.
\newblock {\em ACS central science}, 4(2):268--276, 2018.

\bibitem{irwin2012zinc}
John~J Irwin, Teague Sterling, Michael~M Mysinger, Erin~S Bolstad, and Ryan~G
  Coleman.
\newblock Zinc: a free tool to discover chemistry for biology.
\newblock {\em Journal of chemical information and modeling}, 52(7):1757--1768,
  2012.

\bibitem{kingma2013auto}
Diederik~P. Kingma and Max Welling.
\newblock Auto-encoding variational bayes.
\newblock In {\em 2nd International Conference on Learning Representations,
  {ICLR} 2014, Banff, AB, Canada, April 14-16, 2014, Conference Track
  Proceedings}, 2014.

\bibitem{kusner2017grammar}
Matt~J Kusner, Brooks Paige, and Jos{\'e}~Miguel Hern{\'a}ndez-Lobato.
\newblock Grammar variational autoencoder.
\newblock In {\em International Conference on Machine Learning}, pages
  1945--1954, 2017.

\bibitem{liu2018constrained}
Qi Liu, Miltiadis Allamanis, Marc Brockschmidt, and Alexander Gaunt.
\newblock Constrained graph variational autoencoders for molecule design.
\newblock In {\em Advances in neural information processing systems}, pages
  7795--7804, 2018.

\bibitem{pang2020learning}
Bo Pang, Tian Han, Erik Nijkamp, Song-Chun Zhu, and Ying~Nian Wu.
\newblock Learning latent space energy-based prior model.
\newblock {\em arXiv preprint arXiv:2006.08205}, 2020.

\bibitem{podda2020deep}
Marco Podda, Davide Bacciu, and Alessio Micheli.
\newblock A deep generative model for fragment-based molecule generation.
\newblock {\em arXiv preprint arXiv:2002.12826}, 2020.

\bibitem{samanta2019nevae}
Bidisha Samanta, DE Abir, Gourhari Jana, Pratim~Kumar Chattaraj, Niloy Ganguly,
  and Manuel~Gomez Rodriguez.
\newblock Nevae: A deep generative model for molecular graphs.
\newblock In {\em Proceedings of the AAAI Conference on Artificial
  Intelligence}, volume~33, pages 1110--1117, 2019.

\bibitem{shi2020graphaf}
Chence Shi, Minkai Xu, Zhaocheng Zhu, Weinan Zhang, Ming Zhang, and Jian Tang.
\newblock Graphaf: a flow-based autoregressive model for molecular graph
  generation.
\newblock {\em arXiv preprint arXiv:2001.09382}, 2020.

\bibitem{simonovsky2018graphvae}
Martin Simonovsky and Nikos Komodakis.
\newblock Graphvae: Towards generation of small graphs using variational
  autoencoders.
\newblock In {\em International Conference on Artificial Neural Networks},
  pages 412--422. Springer, 2018.

\bibitem{weininger1988smiles}
David Weininger.
\newblock Smiles, a chemical language and information system. 1. introduction
  to methodology and encoding rules.
\newblock {\em Journal of chemical information and computer sciences},
  28(1):31--36, 1988.

\bibitem{you2018graph}
Jiaxuan You, Bowen Liu, Zhitao Ying, Vijay Pande, and Jure Leskovec.
\newblock Graph convolutional policy network for goal-directed molecular graph
  generation.
\newblock In {\em Advances in neural information processing systems}, pages
  6410--6421, 2018.

\end{thebibliography}

\end{document}